# Deep Learning-assisted AMD Staging based on OCT and OCT Angiography

Yukun Guo[1,2], Tristan T. Hormel[1], An-Lun Wu[1,3], Liqin Gao[1], Min Gao[1,2], Steven T. Bailey[1], Yali Jia[1,2,*]

[1]*Casey Eye Institute, Oregon Health & Science University, Portland, OR 97239, USA*
[2]*Department of Biomedical Engineering, Oregon Health & Science University, Portland, OR 97239, USA*
[3]*Department of Ophthalmology, Mackay Memorial Hospital, Hsinchu 300044, Taiwan*
**jiaya@ohsu.edu*

**Purpose**: To develop deep learning models for automated grading of age-related macular degeneration (AMD) severity using optical coherence tomography (OCT) and OCT angiography (OCTA) data.

**Design**: Cross-sectional study.

**Participants**: Two hundred and seventy-one participants aged ≥50, with a range of AMD severities.

**Methods**: Central macular 6 × 6-mm OCT/OCTA volumes were acquired from all participants enrolled in an AMD study using a swept-source OCTA system (SOLIX; Visionix/Optovue Inc., CA). Each volume consisted of 512 B-frames and each B-frame consisted of 512 A-lines. Each position acquired repeated B-scans for OCTA generation using split-spectrum amplitude-decorrelation angiography (SSADA). AMD severity was graded into four stages (No AMD, Early, Intermediate, Advanced) based on the AREDS simplified severity scale. Three deep learning models were developed using distinct input modalities: (1) biomarker maps derived from segmented pathologic features including retinal fluid, drusen, geographic atrophy (GA), and macular neovascularization (MNV); (2) Two-dimensional (2D) *en face* OCT and OCTA projections of selected tissue layers; and (3) Three-dimensional (3D) OCT/OCTA volumes. Models employed EfficientNet-based architectures (2D EfficientNet-B5 and 3D EfficientNet-B2). Inputs were normalized and resized, and data augmentation (random flopping, scaling, adding Gaussian noise) was applied during training. Five-fold cross-validation to enable robust statistical comparison.

**Main Outcome Measures**: Classification performance was assessed using quadratic weighted kappa (QWK), precision, recall, and F1-score.

**Results**: A total of 2,030 OCT/OCTA volumes from 351 eyes of 271 participants were obtained. All models achieved strong performance in AMD staging, achieving substantial agreement with reference standards (QWK ≥0.83. The biomarker model demonstrated the highest overall performance (QWK, 0.85 ± 0.03, mean±standard deviation) and showed highest accuracy in detecting early AMD (F1-score, 0.59 ± 0.14). The 3D model achieved comparable QWK to the 2D OCT/OCTA model (3D, 0.83 ± 0.04 vs. 2D, 0.83 ± 0.09), while the 2D OCT/OCTA model exhibited the highest precision (0.79 ± 0.06) and most reliably identified No AMD eyes.

**Conclusions**: We proposed the first automated AMD staging system based on OCTA system using deep learning. The models using OCT/OCTA data can accurate and automate AMD staging. The biomarker-based model in this study provided the most balanced and clinically useful performance, particularly for early-stage disease detection.

## Introduction

Age-related macular degeneration (AMD) is a leading cause of vision loss around the world [1,2]. Staging the disease is important for clinical management [3,4]. AMD progression is usually related to specific pathologic features[5–7]. For example, the Age-Related Eye Disease Study (AREDS) grading system is based on biomarker quantification in color fundus photos (CFP) [8,9]. Although AREDS provides a program for staging, it still requires a specialist to use and so is a time-consuming process. To reduce the labor burden that is associated with manual AMD staging an increasing number of studies have proposed automated grading approaches, often leveraging deep learning techniques. Several groups have developed deep learning models that achieve parity with, or even exceed, human AMD diagnostic performance on CFP images [10–12].

While such models are capable of staging AMD, there is reason to believe that these and similar automated procedures could be improved. In particular, CFP imaging has several disadvantages relative to OCT [13], which has become a critical imaging modality ophthalmology because of its ability to capture high-resolution volumetric anatomical information. This capability allows improved visualization of AMD associated retinal lesions such as drusen, retinal fluid, and geographic atrophy (GA). Several previously published works have demonstrated that these lesions can be effectively detected and segmented using deep learning models on OCT data[14–18]. In addition, several studies have explored AMD screening and classification using OCT-based approaches. Venhuizen et.al. proposed a machine learning system to grade OCT data into different AMD severity stages, which demonstrated that a machine learning system can use this modality to achieve similar performance to human graders [19]. However, due to the limited capacity for feature representation, their method exhibits restricted generalizability and is vulnerable to noise. Elsharkawy et al. proposed a complex AI model that derives retinal biomarkers (e.g., fluid, layer disruption, drusen) from OCT and uses them to classify AMD into normal, early, intermediate, geographic atrophy (GA), and active/inactive wet AMD with hierarchical decision logic [20]. However, the AI model only focused on cross-sectional OCT images; no 3D information was involved. Laila et al. used a recurrent deep learning model to classify AMD from OCT images that can classify AMD into normal, dry-AMD, and wet-AMD [21]. Although this study demonstrated the effectiveness of OCT data for AMD staging, most existing OCT-based approaches focus on coarse disease categorization or rely on limited pre-defined biomarkers, which may not fully capture the complex, continuous nature of AMD progression.

OCT can further be augmented with OCTA, an OCT-based data acquisition and processing method that enables detection of vascular function. OCTA provides an extension by generating angiographic information from the volume, enabling assessment of retinal blood flow. OCTA is high-density volumetric imaging modality with micrometer-scale resolution, which can offer increased sensitivity to subtle and early pathological changes that may be missed by conventional or sparsely represented features. This is particularly important for neovascular AMD, where macular neovascularization (MNV) is a critical biomarker [22]. OCTA enables depth-resolved visualization of abnormal flow, allowing more precise detection than structural imaging alone. Despite these advantages, OCTA remains underutilized in automated AMD staging.

This motivates this study, which leverages complementary structural and flow information across biomarker, *en face*, and volumetric representations from combined OCT/OCTA to improve AMD severity classification. Here, we developed a AMD staging framework that integrates both OCT and OCTA information to address key limitations of prior CFP- and OCT-only approaches. Unlike previous OCT-based studies that relied primarily on cross-sectional images or limited *en face* images, our models leverage complementary representations of AMD pathology, including explicitly segmented biomarkers, *en face* projections, and full volumetric data. We implemented three models for four-stage AMD classification and investigated the classification performance of them.

## Methods

### *Data acquisition*

Volumetric OCT/OCTA data were acquired from a 6 × 6-mm central macular region using a swept-source OCTA system operating at 120 kHz (SOLIX; Optovue/Visionix, Inc.). Compared to traditional clinical OCT, this system enables the acquisition of high-density OCT and OCTA volumes simultaneously with OCTA data processed using the split-spectrum amplitude-decorrelation angiography (SSADA) algorithm [22]. In this work two repeat scans in two orthogonal scanning directions were acquired to produce the OCT and OCTA volumes, with each volume registered and motion corrected using a previously described approach [23]. This study was approved by the Institutional Review Board of Oregon Health & Science University (Portland, OR) and adhered to the Declaration of Helsinki.

### *Study Population and Staging Standard Definition*

Participants aged ≥50 years underwent a comprehensive ophthalmic examination, including CFP, and volumetric OCT/OCTA imaging. One or both eyes were included in the examination. AMD severity was staged according to the simplified Age-Related Eye Disease Study (AREDS) scale [8]. CFP served as the primary grading modality for drusen area and pigmentary abnormalities. OCT was used to confirm and characterize structural features, including drusen, retinal fluid, and GA, while OCTA was used to determine the presence of MNV.

Eyes were categorized into four stages. Eyes without AMD ("No AMD") showed no clinical evidence of macular degeneration or other retinal disease (eyes with any other diagnosed retinal disease were excluded). Early AMD eyes demonstrated small drusen (<63 μm) with minimal pigmentary abnormalities. Intermediate AMD eyes exhibited medium-to-large drusen (≥63 μm) and/or more evident pigmentary changes. Advanced AMD eyes included nonexudative disease with geographic atrophy or exudative disease with MNV, confirmed by CFP and OCT/OCTA imaging. Only OCT volumes and corresponding OCTA volumes with signal strength index (SSI) >55 and minimal motion artifacts were included to ensure adequate image quality for analysis. All cases were independently graded by three retina specialists (A.W., L.G., and S.B.), with the final grade determined in consultation with consensus.

A total of 2,030 OCT/OCTA volumes from 351 eyes of 271 participants were obtained. The dataset comprised 147 eyes with No AMD (7.2%), 75 eyes with Early AMD (3.7%), 756 eyes with Intermediate AMD (37.2%), and 1,052 eyes with Advanced AMD (51.8%).

### *Deep learning model design*

In this study, we developed three deep learning models for automated AMD staging using OCT and OCTA data. The biomarker model (Fig. 1A) used a 2D CNN with EfficientNet-B5 [24] backbone to extract features from four segmented AMD-related biomarker maps (retinal fluid, drusen, GA, and MNV). The 2D OCT/OCTA model (Fig. 1B) shared the same 2D CNN architecture with four *en face* OCT/OCTA images as input to capture structural and vascular features for classification. The 3D OCT/OCTA model (Fig. 1C) employed a 3D CNN with an EfficientNet-B2 backbone to directly learn from full volumetric OCT and OCTA data. EfficientNet backbones were chosen to balance feature extraction performance with computational efficiency.

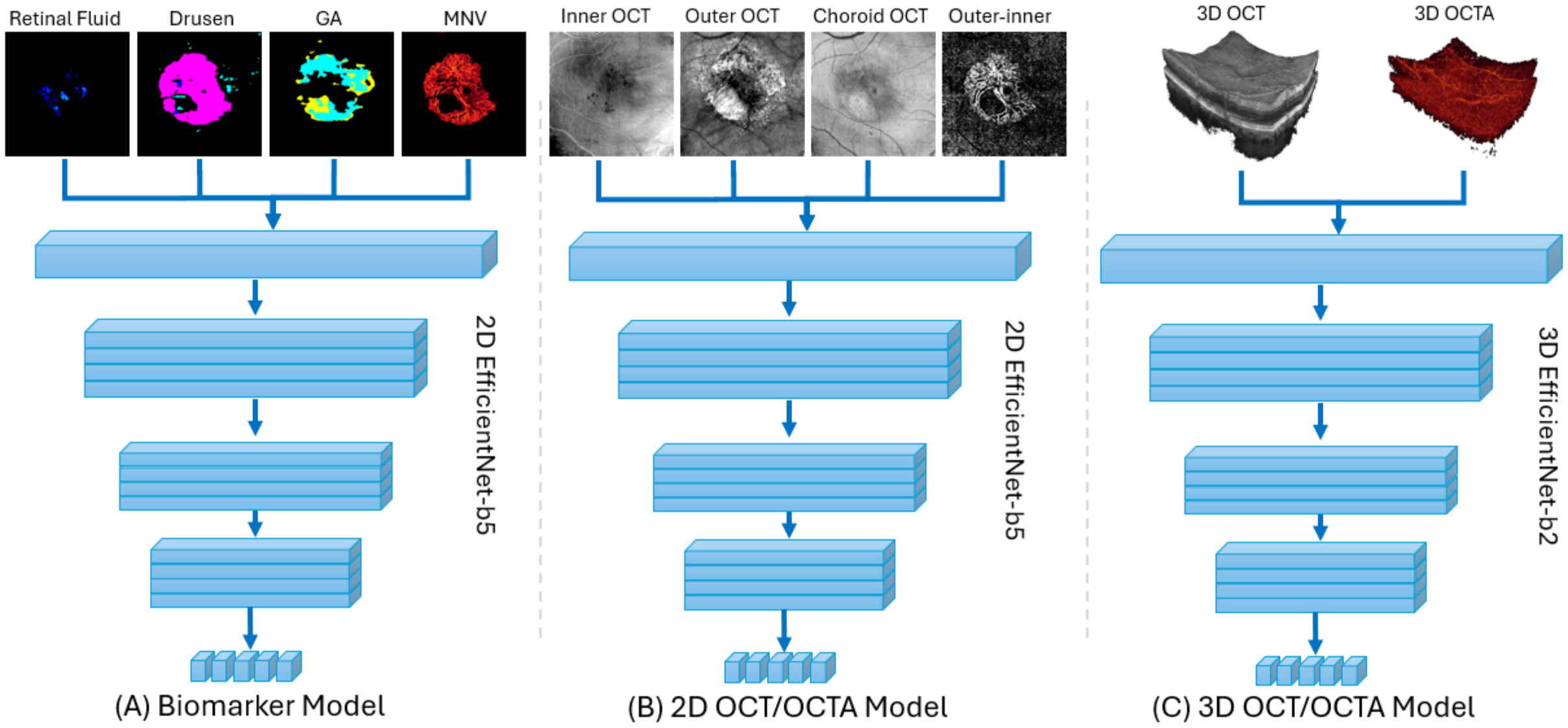

Figure 1. Overview of the deep learning architectures for automated AMD staging. (A) Biomarker model. The input includes four segmented biomarkers: retinal fluid, drusen, geographic atrophy (GA), and macular neovascularization (MNV). (B) 2D OCT/OCTA model. The input includes four *en face* OCT images: inner retinal OCT, choroidal OCT, outer retinal OCT, and a subtraction-based OCTA image derived from outer and inner retinal slabs. (C) 3D OCT/OCTA model. The input includes the final grade determined in consultation with consensus.

***Preprocessing***

The input data for the Biomarker model [Fig. 1 A] consists of four retinal biomarkers that are closely associated with AMD severity: retinal fluid, drusen, GA, and MNV. Retinal fluid [Fig. 2 A] was segmented by our previously published deep learning model [14]. Drusen [Fig. 2 B] information was derived from images that included four distinct subtypes. These subtypes were segmented and classified based on our prior work [15], in which we proposed a hybrid approach combining a deep learning segmentation network with a subsequent classification algorithm to accurately identify and differentiate from OCT volumes. The GA [Fig. 2 C] maps were generated using a deep learning model that was developed in our previous work [25]. The MNV map [Fig. 2D] was generated by extracting the OCTA flow signal from the OCTA volume within the segmented drusenoid PED/MNV regions [15], followed by maximum intensity projection to produce the corresponding *en face* OCTA.

For the 2D OCT/OCTA model [Fig. 1 B], the input includes four 2D *en face* OCT/OCTA images. These *en face* images are generated by projecting anatomically relevant slabs from the OCT/OCTA volumes that correspond to the four retinal lesions used in the biomarker model: The *en face* OCT of the inner retina [Fig. 2 E], corresponding to retinal fluid, is generated from the slab extending from the inner limiting membrane (ILM) to the outer nuclear layer (ONL). The *en face* OCT of outer retina [Fig. 2 F], corresponding to drusen, is generated by projecting the slab between the ellipsoid zone (EZ) and Bruch's membrane (BM). The *en face* OCT of the choroid [Fig. 2 G], corresponding to GA, is generated by projecting the choroidal slab, which captures signal alterations associated with atrophic changes. The *en face* OCTA of MNV [Fig. 2 H], is generated by subtracting the inner retinal OCTA projection from the outer retinal OCTA projection. This subtraction suppresses projection artifacts originating from superficial retinal blood flow and enhances the visualization of neovascular signals in the outer retina.

For the 3D OCT/OCTA model [Fig. 1 C], the input consists of OCT and OCTA volumes, enabling the network to leverage full 3D structural and vascular information for lesion characterization and AMD severity assessment.

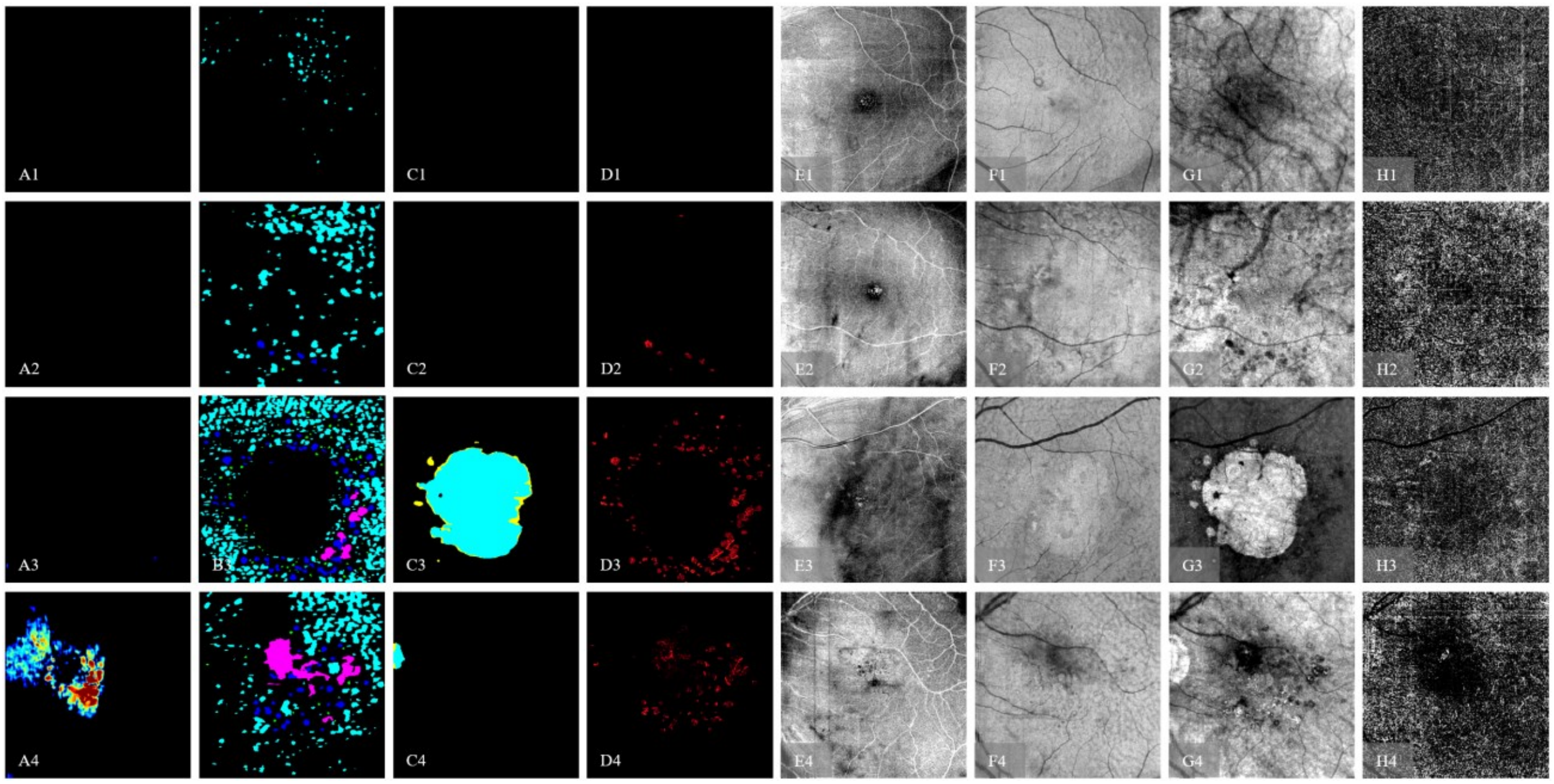

Figure 2. Representative examples of input data for the Biomarker model and the 2D OCT/OCTA model across AMD stages. (A) *En face* retinal fluid heat maps; warmer colors correspond to areas of increased fluid accumulation. (B) Drusen subtype maps including four categories: cyan, pseudodrusen; green, small drusen; blue, large drusen; and magenta, drusenoid pigment epithelial detachment (PED). (C) Geographic atrophy (GA) maps, where cyan denotes complete retinal pigment epithelium and outer retinal atrophy(cRORA), and yellow indicates choroidal hyper-transmission defects. (D) Macular neovascularization (MNV) maps. (E) *En face* OCT of inner retina. (F) *En face* OCT of outer retina. (G) *En face* OCT of choroid. (H) *En face* OCTA obtained by subtracting the inner retinal slab from the outer retinal slab. Rows correspond to increasing AMD severity: the first row shows an eye with early AMD; the second row shows intermediate AMD; the third row shows advanced dry AMD; and the fourth row shows advanced AMD with retinal fluid.

### ***Training Configuration and Hyperparameters***

#### *Loss Function and Optimizer*

The model was trained using categorical cross-entropy loss to optimize multi-class classification performance. The Adam optimizer was employed with an initial learning rate of 0.001. A learning rate scheduler was implemented to dynamically adjust the learning rate by a factor of 0.1 when validation loss plateaued for 3 consecutive epochs, with a minimum learning rate threshold of $1\times10^{-8}$ to maintain training stability.

#### *Data Augmentation*

For training, data augmentation was used, including the addition of random Gaussian noise, random horizontal flipping, and scaling (304×304 for 2D models, 192×256×256 for 3D models) from the original size (640×512×512). Data was normalized to [0, 1] before feed into network for training. No augmentation was applied to the validation and test datasets to ensure unbiased evaluation.

#### *Cross-Validation*

The entire dataset was first stratified into a training set (80%) and a hold-out test set (20%), while preserving the original class distribution. To prevent data leakage, all volumes acquired from the same eye were assigned to only one subset. Five-fold cross-validation was adopted to ensure robust model evaluation. The cross-validation was only performed on the training set.  After completing training within each fold, the optimized model was evaluated on the hold-out test set. Performance metrics were averaged across the five training iterations to obtain the final results.

All models were trained for a maximum of 500 epochs with a batch size of 16. Early stopping was implemented with a patience of 10 epochs, monitoring validation loss to prevent overfitting and reduce unnecessary computational cost. All experiments were implemented in PyTorch (version 2.9.1) using Python 3.11 and CUDA 12.6 on a Linux-based workstation equipped with two NVIDIA RTX 3090 GPUs (24 GB RAM each) and 512 GB system memory. Multi-GPU was employed to accelerate training. Mixed-precision training was implemented using PyTorch Automatic Mixed Precision with dynamic loss scaling to improve computational efficiency and reduce memory consumption.

*Evaluation Methods*

Model performance was evaluated using multiple complementary metrics. All metrics were calculated using macro-averaging across the four AMD severity stages. This two-stage aggregation approach ensures equal weighting of all disease classes regardless of their prevalence in the dataset, which is critical given the substantial class imbalance in our data.

*Performance Metrics*

For each AMD class $c$ and fold $k$, the following metrics were computed from the confusion matrix elements: $\mathrm{TP}_{c,k}$, True Positives: correctly predicted as class $c$. $\mathrm{TN}_{c,k}$ , True Negatives: correctly predicted as not class $c$. $\mathrm{FP}_{c,k}$ , False Positives: incorrectly predicted as class $c$, and $\mathrm{FN}_{c,k}$ , False Negatives: incorrectly predicted as not class $c$ when the true class is $c$. The Recall, Precision, and F1-score are calculated via the following equations based on the confusion matrix elements.

$$\mathrm{Recall}_{c,k} = \frac{\mathrm{TP}_{c,k}}{\mathrm{TP}_{c,k} + \mathrm{FN}_{c,k}} \tag{1}$$

$$\mathrm{Precision}_{c,k} = \frac{TP_{c,k}}{\mathrm{TP}_{c,k} + \mathrm{FP}_{c,k}} \tag{2}$$

$$\mathrm{F1\ score}_{c,k} = \frac{2 \cdot \mathrm{TP}_{c,k}}{2 \cdot \mathrm{TP}_{c,k} + \mathrm{FP}_{c,k} + \mathrm{FN}_{c,k}} \tag{3}$$

We also include the quadratic weighted kappa (QWK), which measures agreement between predicted and true labels while accounting for the ordinal nature of AMD severity stages and weighting disagreements quadratically by their distance:

$$\mathrm{QWK}_k = 1 - \frac{\sum_{i=1}^{4}\sum_{j=1}^{4} w_{i,j} O_{i,j}}{\sum_{i=1}^{4}\sum_{j=1}^{4} w_{i,j} E_{i,j}} \tag{4}$$

where $O_{i,j}$ is the observed count of samples with true class $i$ predicted as class $j$ (from the confusion matrix), $E_{i,j}$ is the expected count under chance agreement: $E_{i,j} = (n_i \times m_j)/N$, with $n_i$= total samples of true class $i$, $m_j$= total predictions as class $j$, and $N$ = total samples. $w_{i,j}$is the quadratic weight: $w_{i,j}$= $(i-j)^2/(C-1)^2$, where $C$ = 4 (number of classes). QWK ranges from -1 (complete disagreement) to 1 (perfect agreement), with 0 indicating chance-level agreement. The quadratic weighting scheme penalizes large classification errors while being more tolerant of adjacent misclassifications. This makes QWK particularly appropriate for ordinal AMD staging, where clinical consequences of misclassification increase with distance between predicted and true severity levels.

*Macro-Averaging*

For each cross-validation fold $k$ (where $k$ = 1, 2, ..., 5), the macro-averaging was calculated through the following procedure. First, for each AMD severity class $c$, compute the class-specific metric $M_{c,k}$ using the test set predictions from fold $k$. Second, compute the fold level macro-average metric as:

$$M_k^{\text{macro}} = \frac{1}{4}\sum_{c=1}^{4} M_{c,k} \tag{5}$$

where the sum is over all four AMD classes, ensuring equal contribution from each class regardless of sample size. Then compute the overall mean and standard deviation across all 5 folds.

## Results

### *Overall Classification Performance*

The classification performance of the three models was evaluated using five-fold cross-validation [Table 1]. All models demonstrated strong and consistent performance across AMD severity, with substantial agreement to reference labels. Among them, the Biomarker model achieved the best overall performance and showed the most stable results across cross-validation folds. The 2D OCT/OCTA model showed comparable performance, indicating that *en face* structural and angiographic information provides robust classification power. The 3D OCT/OCTA model also achieved high discriminative performance but exhibited greater variability across folds.

**Table 1. Overall Performance Comparison of Three Models (Macro-Averaging)**

| Model | Recall | Precision | F1-Score | QWK |
|---|---|---|---|---|
| **Biomarker** | 0.78 ± 0.05 | 0.76 ± 0.06 | 0.77 ± 0.05 | 0.85 ± 0.03 |
| **2D OCT/OCTA** | 0.73 ± 0.07 | 0.79 ± 0.06 | 0.74 ± 0.06 | 0.83 ± 0.04 |
| **3D OCT/OCTA** | 0.71 ± 0.14 | 0.74 ± 0.13 | 0.72 ± 0.14 | 0.83 ± 0.09 |

*Note: Values represent mean ± standard deviation across 5 folds. QWK: Quadratic Weighted Kappa.*

All three models achieved precision values ranging from 0.74 to 0.79, demonstrating the low false-positive classification that are important for avoiding unnecessary clinical burden. The strong recall values achieved across all three models (0.71–0.78), indicating that the models reliably detect the majority of true AMD cases at each severity stage. Furthermore, QWK values between 0.83 and 0.85 indicate substantial agreement with the reference labels.

### *Per-Class Performance Analysis*

Across all models, per-class performance varied by AMD severity [Table 2]. For all models, No AMD eyes were classified with high accuracy, with F1-scores ranging from 0.82 to 0.92; the 2D OCT/OCTA model achieved the highest F1-score (0.92 ± 0.03) and recall (0.92 ± 0.06). Early AMD showed reduced performance across models, with the biomarker model achieving the highest F1-score (0.59 ± 0.14) and recall (0.65 ± 0.17), followed by the 2D OCT/OCTA and 3D OCT/OCTA models. Performance improved for Intermediate AMD, with F1-scores between 0.76 and 0.79; the 3D OCT/OCTA model achieved the highest recall (0.83 ± 0.06) and F1-score (0.79 ± 0.08). Advanced AMD was classified reliably by all models, with F1-scores ranging from 0.84 to 0.88; the biomarker and 3D OCT/OCTA models achieved the highest F1-scores (0.88), with balanced recall and precision (0.87–0.89).

**Table 2. Per-Class Performance Metrics for Each Model (mean ± standard deviation)**

| Class | Model | Recall | Precision | F1-Score |
|---|---|---|---|---|
| No AMD | **Biomarker** | 0.84 ± 0.05 | 0.81 ± 0.04 | 0.82 ± 0.04 |
| | **2D OCT/OCTA** | 0.92 ± 0.06 | 0.92 ± 0.07 | 0.92 ± 0.03 |
| | **3D OCT/OCTA** | 0.80 ± 0.17 | 0.90 ± 0.05 | 0.84 ± 0.12 |
| Early AMD | **Biomarker** | 0.65 ± 0.17 | 0.56 ± 0.16 | 0.59 ± 0.14 |
| | **2D OCT/OCTA** | 0.38 ± 0.20 | 0.64 ± 0.18 | 0.46 ± 0.20 |
| | **3D OCT/OCTA** | 0.35 ± 0.35 | 0.41 ± 0.39 | 0.36 ± 0.34 |
| Intermediate AMD | **Biomarker** | 0.76 ± 0.04 | 0.80 ± 0.03 | 0.78 ± 0.03 |
| | **2D OCT/OCTA** | 0.82 ± 0.02 | 0.71 ± 0.04 | 0.76 ± 0.02 |
| | **3D OCT/OCTA** | 0.83 ± 0.06 | 0.76 ± 0.11 | 0.79 ± 0.08 |
| Advanced AMD | **Biomarker** | 0.89 ± 0.03 | 0.87 ± 0.02 | 0.88 ± 0.02 |
| | **2D OCT/OCTA** | 0.80 ± 0.04 | 0.88 ± 0.02 | 0.84 ± 0.02 |
| | **3D OCT/OCTA** | 0.87 ± 0.05 | 0.89 ± 0.05 | 0.88 ± 0.05 |

*Note: Values represent mean ± standard deviation across 5 folds.*

*Confusion Matrix Analysis*

Figure 3 presents the normalized confusion matrices for the three models across four AMD severity stages. Misclassifications primarily occurred between adjacent stages (e.g., No AMD vs. Early AMD, Early AMD vs. Intermediate AMD). High correct classification rates were observed for No AMD (80–92%), Intermediate AMD (76–83%), and Advanced AMD (80–89%), whereas Early AMD showed lower accuracy (35–65%) across models. Early AMD was frequently misclassified as No AMD or Intermediate AMD; the 2D OCT/OCTA model achieved 38% correct classification for Early AMD, while the biomarker model showed higher accuracy (65%) with reduced off-diagonal errors. Overall, the biomarker model demonstrated comparatively higher diagonal values, while the 3D OCT/OCTA model showed greater misclassification for Early AMD but maintained strong performance for the remaining classes.

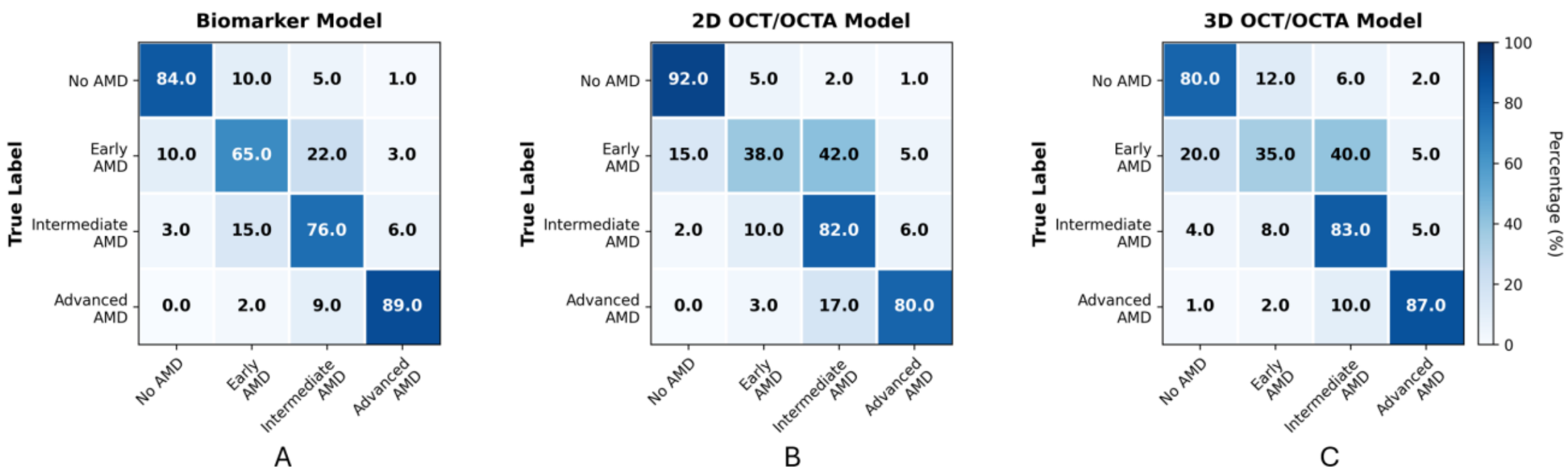

Figure 3. Normalized confusion matrixes for AMD severity staging. (A) Biomarker model. (B) 2D OCT/OCTA model. (C) 3D OCT/OCTA model.

Across the five cross-validations, the biomarker model showed the narrowest distributions across most metrics, indicating superior consistency compared to other models (Fig. 4). The 2D OCT/OCTA model demonstrated moderate variability, whereas the 3D OCT/OCTA model exhibited wider distributions, particularly for F1-score and recall. Precision displayed minimal variability for all models, while F1-score and recall varied more across folds. No outliers were observed, and the mean and median values were closely aligned, indicating stable performance distributions.

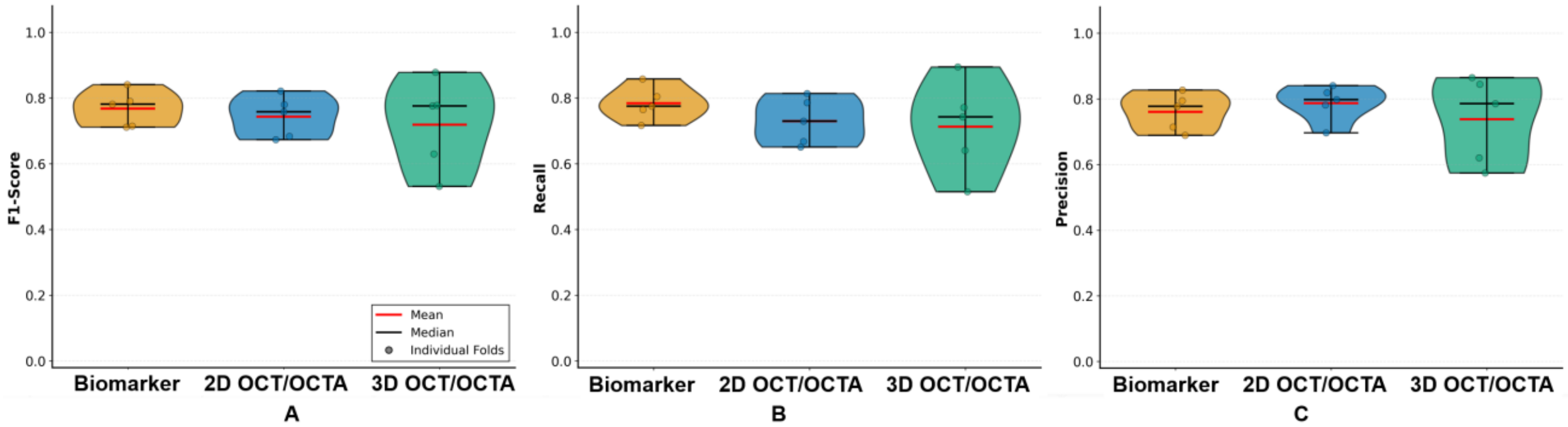

Figure 4. Performance Distribution Across Five-Fold Cross-Validation. Plots showing distribution of (A) F1-Score, (B) Recall, and (C) Precision.

## Discussion

In this study, we developed and evaluated three deep learning models for automated AMD severity staging using combined OCT and OCTA data, demonstrating high classification performance (QWK ≥ 0.83) across four clinically relevant stages (No AMD, Early AMD, Intermediate AMD, and Advanced AMD). Existing automated AMD staging approaches have relied predominantly on CFP or traditional OCT, each carrying notable limitations. CFP-based approaches are constrained by the modality's limited depth resolution and inability to detect vascular pathology such as MNV [10–12]. OCT systems, while superior to CFP in anatomical detail, similarly lack direct access to retinal blood flow information and have been developed from cross-sectional or sparsely sampled volumes that do not fully capture the spatial complexity of AMD pathology. Our framework addresses these gaps by leveraging swept-source OCTA, which simultaneously acquires high-density, co-registered OCT and OCTA volumes in a single non-invasive scan. This combination of imaging modalities enables simultaneous extraction of AMD staging related pathological features, including drusen, retinal fluid, geographic atrophy, and MNV. By incorporating this information, our approach provides a more comprehensive and clinically informative basis for automated AMD grading than prior methods.

Among the three architectures evaluated, each offered distinct performance characteristics that reflect both the nature of their inputs and the inherent challenges of the AMD staging task. The Biomarker model, which inputs the *en face* projections of four segmented pathological features, achieved the most consistent overall performance, with the highest recall (0.78±0.05), F1-score (0.77±0.05) and QWK (0.85±0.03), as well as lowest variability in cross-validation [Fig. 4]. Its advantage was most pronounced in Early AMD detection, where it achieved a recall of 0.65 with 71% higher than the 2D OCT/OCTA model (0.38) and nearly double that of the 3D OCT/OCTA model. This superior performance can be attributed to several factors. First, the explicit segmentation and representation of disease-specific biomarkers provide the model with stable and diagnostically relevant features that align directly with clinical grading criteria. Second, by isolating individual pathological features, the biomarker approach may reduce the confounding effects of inter-patient anatomical variability and imaging artifacts. Third, the OCTA-derived vascular marker, MNV, offers critical pathophysiological specificity that helps distinguish advanced stages where non-vascular tissue changes alone may be subtle or ambiguous. The consistency of the Biomarker model's performance across cross-validation folds (lowest standard deviations: F1-Score ±0.05, QWK ±0.03) suggests that biomarker-based representations are more robust to variations in training data composition compared to raw imaging features. In addition, reliance on biomarkers makes it more inherently interpretable than most deep learning approaches.

The 2D OCT/OCTA model, which processes four anatomically selected *en face* projections, offered a pragmatic middle ground. It achieved the highest precision (0.79 ± 0.06) and the best No AMD detection performance (F1-score 0.92 ± 0.03), reflecting the discriminability of *en face* structural and angiographic features for identifying healthy retina. However, its substantially lower Early AMD recall (0.38) relative to the Biomarker model underscores the difficulty of detecting subtle early-stage pathology from projection images alone and highlights the value of explicit biomarker pre-processing as a form of guided feature extraction.

Contrary to the theoretical expectation that volumetric inputs would yield superior performance, the 3D OCT/OCTA exhibited the highest performance variability (F1-Score ±0.14, QWK ±0.09) and did not consistently outperform 2D approaches. This finding reflects several challenges inherent to 3D medical image analysis with limited training data. First, 3D convolutional architecture requires substantially more parameters than their 2D counterparts, increasing the risk of overfitting when training data is limited. Second, our training cohort of 1,695 samples, while substantial for 2D analysis, may be insufficient to fully exploit the information capacity of 3D volumetric representations; this was particularly problematic for the small subset of Early AMD in this study (a fact reflected in in the 3D model's performance identifying this severity). Third, the full 3D OCT/OCTA data may contain substantial noise and irrelevant information that increase the challenge for 3D model to extract critical features for classification.

Despite the strong results, there are several limitations in this study. First, Early AMD detection remained the most challenging task across all models, with recall ranging from 0.35 to 0.65. Contributing factors include severe class imbalance, with only 64 Early AMD cases (3.8%) versus 880 Advanced AMD cases (51.9%), leading models to favor majority classes during training. Early AMD also exhibits subtle and heterogeneous morphological changes, resulting in intrinsic diagnostic ambiguity even among expert graders. Confusion predominantly occurred with adjacent grades (Early AMD and Intermediate AMD), reflecting the continuous progression of AMD and highlighting the need for finer discrimination along the severity continuum. Second, the biomarker-based framework depends on upstream biomarker segmentation models, and classification performance may be influenced by segmentation accuracy. While these segmentation algorithms were previously validated and strict image quality criteria were applied, error propagation was not quantified in this study. Third, external validation on independent cohorts was not performed. Although five-fold cross-validation demonstrated consistent internal performance, evaluation on outside datasets is necessary to confirm generalizability.

In future work, we plan to expand the cohort with increased representation of early-stage AMD to mitigate class imbalance and improve minority-class recall. We also aim to validate the models on larger, independent datasets to further enhance robustness and generalizability.

## Conclusion

Our study demonstrates that deep learning models using OCT/OCTA data from OCTA system can achieve clinically relevant automated AMD staging with high overall performance (QWK ≥0.83). The high-density retinal imaging provided by OCTA system enables deep learning models to extract essential structural and angiographic features for accurate classification. The biomarker-based model showed superior and more consistent performance across all stages, particularly for Early AMD. With further refinement and rigorous validation, AI-assisted AMD staging has strong potential to improve early detection, standardize monitoring, and enhance access to timely interventions in clinical practice.

## Founding

This work was supported by grants from National Institutes of Health (R01 EY 036429, R01 EY035410, R01 EY024544, R01 EY027833, R01 EY031394, R43EY036781, P30 EY010572, T32 EY023211, UL1TR002369); the Jennie P. Weeks Endowed Fund; the Malcolm M. Marquis, MD Endowed Fund for Innovation; Unrestricted Departmental Funding Grant and Dr. H. James and Carole Free Catalyst Award from Research to Prevent Blindness (New York, NY); Edward N. & Della L. Thome Memorial Foundation Award, and the Bright Focus Foundation (G2020168, M20230081).

### Disclosures

Yukun Guo: Optovue/Visionix (P), Genentech/Roche (P, R); Ifocus Imaging (P), Tristan T. Hormel: Ifocus Imaging (I), Tristan T. Hormel: Ifocus Imaging (I), Steven Baily: Visionix/Optovue (F), Yali Jia: Optovue/Visionix (P, R), Genentech/Roche (P, R, F), Ifocus Imaging (I, P), Optos (P), Boeringer Ingelheim (C), Kugler (R)

### Data availability

Data underlying the results presented in this paper are not publicly available at this time but may be obtained from the authors upon reasonable request.